\def\year{2020}\relax
\documentclass[letterpaper]{article}
\usepackage{aaai20}
\usepackage{times}
\usepackage{helvet}
\usepackage{courier}
\usepackage[hyphens]{url}
\usepackage{graphicx}
\usepackage{graphicx}
\title{Pose-Guided Multi-Granularity Attention Network \\for Text-Based Person Search}
\author{Ya Jing,\textsuperscript{\rm 1,3}
Chenyang Si,\textsuperscript{\rm 1,3}
Junbo Wang,\textsuperscript{\rm 1,3}
Wei Wang,\textsuperscript{\rm 1,3}\protect\thanks{Corresponding Author: Wei Wang}
Liang Wang,\textsuperscript{\rm 1,2,3}
Tieniu Tan\textsuperscript{\rm 1,2,3}\\
\textsuperscript{\rm 1}Center for Research on Intelligent Perception and Computing (CRIPAC),\\
National Laboratory of Pattern Recognition (NLPR)\\
\textsuperscript{\rm 2}Center for Excellence in Brain Science and Intelligence Technology (CEBSIT),\\
Institute of Automation, Chinese Academy of Sciences (CASIA)\\
\textsuperscript{\rm 3}University of Chinese Academy of Sciences (UCAS)\\
\{ya.jing, chenyang.si, junbo.wang\}@cripac.ia.ac.cn,
\{wangwei, wangliang, tnt\}@nlpr.ia.ac.cn
}

\begin{document}

\maketitle

\begin{abstract}
  Text-based person search aims to retrieve the corresponding person images in an image database by virtue of a describing sentence about the person, which poses great potential for various applications such as video surveillance. Extracting visual contents corresponding to the human description is the key to this cross-modal matching problem. Moreover, correlated images and descriptions involve different granularities of semantic relevance, which is usually ignored in previous methods. To exploit the multilevel corresponding visual contents, we propose a pose-guided multi-granularity attention network (PMA). Firstly, we propose a coarse alignment network (CA) to select the related image regions to the global description by a similarity-based attention. To further capture the phrase-related visual body part, a fine-grained alignment network (FA) is proposed, which employs pose information to learn latent semantic alignment between visual body part and textual noun phrase. To verify the effectiveness of our model, we perform extensive experiments on the CUHK Person Description Dataset (CUHK-PEDES) which is currently the only available dataset for text-based person search. Experimental results show that our approach outperforms the state-of-the-art methods by 15 \% in terms of the top-1 metric.
\end{abstract}

\section{Introduction}

Person search has gained great attention in recent years due to its wide applications in video surveillance, e.g., missing persons searching and suspects tracking. With the explosive increase in the number of videos, manual person search in large-scale videos is unrealistic, so we need to design automatic methods to perform this task more efficiently. Existing methods of person search are mainly classified into three categories according to the query type, e.g., image-based query \cite{zhou2018graph,Zheng2011Person,Sarfraz2017A}, attribute-based query \cite{Su2016Deep,Vaquero2009Attribute} and text-based query \cite{Li2017Person,Li2017Identity,zheng2017dual,Chen2018Improving}. Image-based person search needs at least one image of the queried person which in many cases is very difficult to obtain. Attribute-based person search has limited capability of describing persons' appearance. Since textual descriptions are more accessible and can describe persons of interest with more details in a more natural way, we study the task of text-based person search in this paper.

\begin{figure}[t]
\centering
\includegraphics[width=0.8\columnwidth]{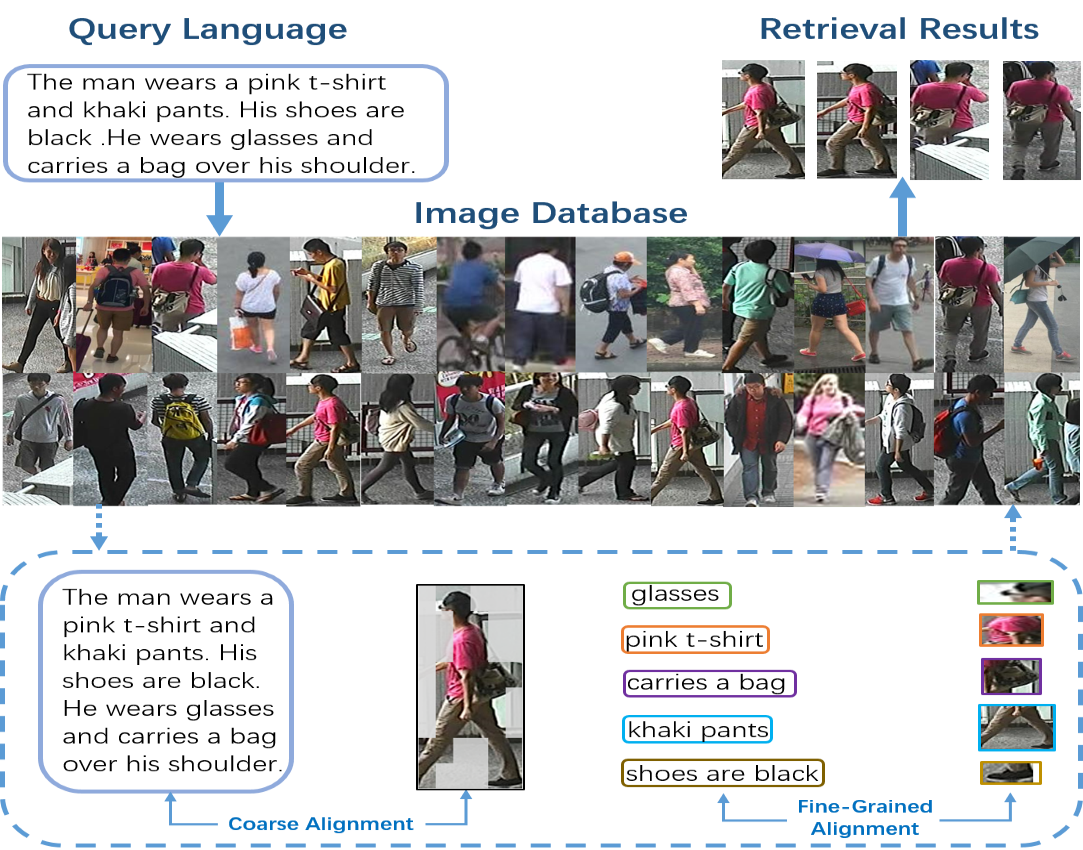}
\caption{The illustration of text-based person search. Given a textual description of a person, the model aims to retrieve the corresponding person images from the given image database. There are two granularity alignments in our retrieval procedure: coarse alignment by textual description and fine-grained alignment by noun phrases.
}
\label{fig:graph1}
\end{figure}

Text-based person search aims to retrieve the corresponding person images to a textual description from a large-scale image database, which is illustrated in Fig.~\ref{fig:graph1}. The main challenge of this task is to effectively extract corresponding visual contents to the human description under multi-granularity semantic relevances between image and text.
Previous methods \cite{zheng2017dual,zhang2018deep} generally utilize Convolutional Neural Networks (CNNs) \cite{krizhevsky2012imagenet} to obtain a global representation of the input image, which often cannot effectively extract visual contents corresponding to the person in image.
Considering that human pose is closely related to human body parts, we exploit pose information for effective visual feature extraction. To our knowledge, we are probably the first to employ human pose to handle the task of text-based person search.

Nonetheless, it is still not suitable to match the global features directly due to the fact that only partial image regions are corresponding to the given textual description. The learned global visual representation suffers from much irrelevant information brought by useless visual regions, e.g., meaningless background. Therefore, selecting description-related image regions is necessary for better matching.
Previous methods \cite{Chen2018Improving,chen2018improve} mainly focus on utilizing attention mechanisms for single-level alignment,
which largely neglect the multi-granularity semantic relevances.
In contrast, we propose to exploit the multi-granularity corresponding visual contents by the given textual description. First, we coarsely select the related visual regions by the whole text.
In addition to sentence-level correlations, phrase-level relations are also significant for fine-grained image-text matching. In fact, the noun phrase in textual description usually is related to the specific visual human part but not the whole image as shown in Fig.~\ref{fig:graph1}. To select the phrase-level visual contents, we learn the latent semantic alignment between noun phrase and visual human part.
Considering that human pose is closely related to human body parts, we utilize pose information to guide the fine-grained alignment.

\begin{figure*}[t]
\centering
\includegraphics[width=0.85\textwidth]{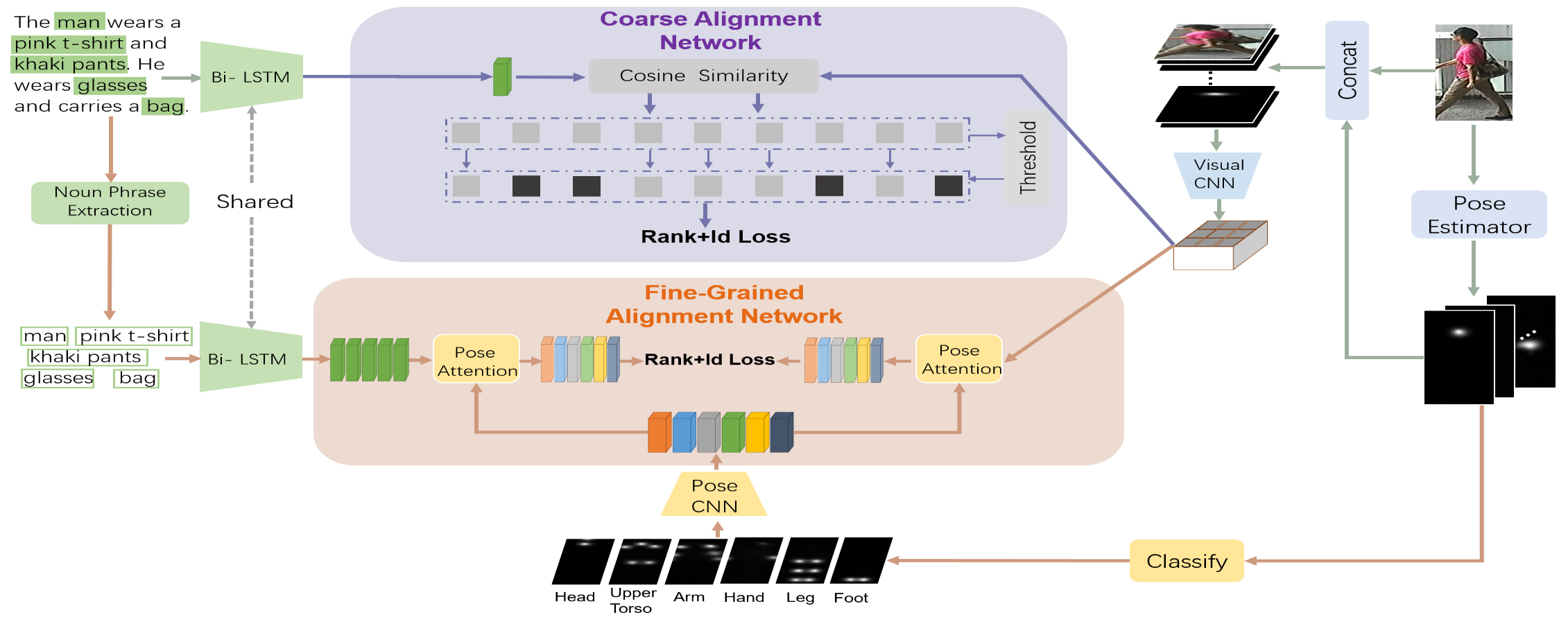}
\caption{The architecture of our proposed pose-guided multi-granularity attention network (PMA) for text-based person search. It contains two matching networks (best viewed in colors): the coarse alignment network aims to select the description-related image regions by similarity-based attention and fine-grained alignment network aims to select phrase-related visual human parts by pose-guided attention.
To better train PMA, both ranking loss and identification loss are employed in our method.
}
\label{fig:model1}
\end{figure*}

In summary, we propose a pose-guided multi-granularity attention network (PMA) for text-based person search as shown in Fig.~\ref{fig:model1}.
First, we estimate human pose from the input image,
where global visual representations emphasize the human-related features by concatenating the pose confidence maps to the original input image. To further capture the description-related image regions, a coarse alignment network (CA) is performed between textual description and image regions, where a similarity-based attention is utilized. To exploit the phrase-related visual contents, a fine-grained alignment network (FA) is further proposed, where the pose information is utilized to guide the attention of noun phrases and image regions.
Both ranking loss and identification loss are employed for better training. Our proposed method is evaluated on a challenging dataset CUHK-PEDES \cite{Li2017Person}, which is currently the only available dataset for text-based person search. Experimental results show that our PMA outperforms the state-of-the-art methods on this dataset.

The main contributions of this paper can be summarized as follows: (1) Multi-granularity corresponding visual contents are extracted by the coarse alignment network and fine-grained alignment network.
(2) A novel pose-guided part aligned matching network is proposed to exploit the latent semantic alignment between visual body part and textual noun phrase, which is probably the first used in text-based person search.
(3) The proposed PMA achieves the best results on the challenging dataset CUHK-PEDES. Extensive ablation studies verify the effectiveness of each component in the PMA.

\section{Related Work}
In this section, we briefly introduce the related work about prior studies on text-based person search, human pose for person search, as well as attention for person search.

\textbf{Text-Based Person Search.} Li et al. \cite{Li2017Person} propose the task of text-based person search and further employ a recurrent neural network with gated neural attention (GNA-RNN) for this task. To utilize identity-level annotations, Li et al. \cite{Li2017Identity} propose an identity-aware two-stage framework.
PWM+ATH \cite{Chen2018Improving} utilizes a word-image patch matching model in order to capture the local similarity. Different from the above three methods which are all the CNN-RNN architectures, Dual Path \cite{zheng2017dual} employs CNN for textual feature learning. CMPM+CMPC \cite{zhang2018deep} utilizes a cross-modal projection matching (CMPM) loss and a cross-modal projection classification (CMPC) loss to learn discriminative image-text representations.
In this paper, we propose a pose-guided multi-granularity attention network to learn multilevel cross-modal relevances.

\textbf{Human Pose for Person Search.} With the development of pose estimation \cite{Insafutdinov2016DeeperCut,cao2016realtime,Carreira2015Human,Chu2016Structured,chu2017multi}, many approaches in image-based person search extract human poses to improve the visual representation. To solve the problem of human pose variations, Pose-transfer \cite{liu2018pose} employs a pose transferrable person search framework through pose-transferred sample augmentations. PIE \cite{zheng2017pose} utilizes human pose to normalize the person image. The normalized image and original image are both used to match the person. SSDAL \cite{Su2016Deep} also utilizes pose to normalize the person image, while it leverages human body parts and the global image to learn a robust feature representation. PSE \cite{Sarfraz2017A} directly concatenates pose information to the input image to learn visual representation. In this paper, we use pose information for effective visual feature extraction and aligned part matching between noun phrase and image regions.

\textbf{Attention for Person Search.} Attention mechanism aims to select key parts of an input, which is generally divided into soft attention and hard attention. Soft attention computes a weight map and selects the input according to the weight map, while hard attention just preserves one or a few parts of the input and ignores the others. Recently, attention is widely used in person search, which selects either visual contents or textual information.
GNA-RNN \cite{Li2017Person} computes an attention map based on text representation to focus on visual units. IATV \cite{Li2017Identity} employs a co-attention method, which extracts word-image features via spatial attention and aligns sentence structures via a latent semantic attention. Zhao et al. \cite{Zhao2017Deeply} employ human body parts to weight the image feature map. As we know, hard attention is rarely exploited in person search. In this paper, we fully exploit the hard and soft attentions in learning the semantic relevance between image and text.

\section{Pose-Guided Multi-Granularity Attention Network }
In this section, we explain the proposed pose-guided multi-granularity attention network in detail. First, we introduce the procedures of visual and textual representations extraction. Then, we describe the two alignment networks including coarse alignment network and fine-grained alignment network. Finally, we give the details of learning the proposed model.

\subsection{Visual Representation Extraction}
Considering that human pose is closely related to human body parts, we exploit pose information to learn human-related visual feature and guide the aligned part matching. In this work, we estimate human pose from the input image using the PAF approach proposed in \cite{Cao2017Realtime} due to its high accuracy and realtime performance.
To obtain more accurate human poses, we retrain PAF on a larger AI challenge
dataset \cite{Wu2017AI} which annotates 14 keypoints for each person.

\begin{figure}[t]
\centering
\includegraphics[width=0.9\columnwidth]{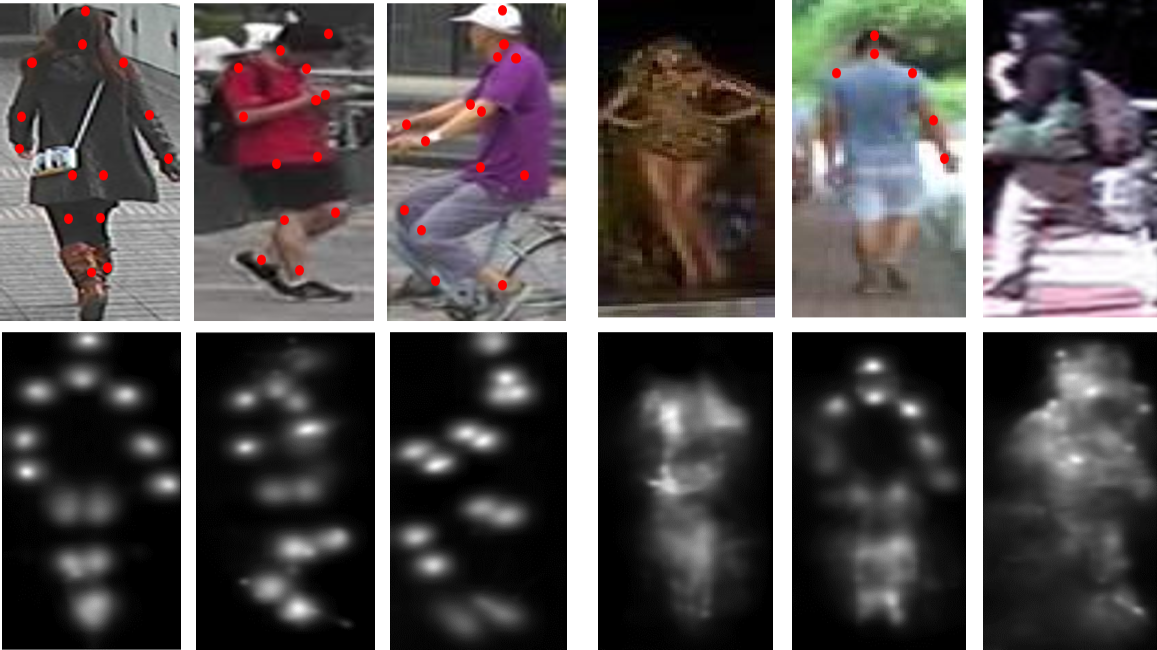}
\caption{Examples of human pose estimation from the retrained PAF. The first row shows the detected joint keypoints and the second row shows the superimposed results from 14 confidence maps.}
\label{fig:pose}
\end{figure}

However, in the experiments, the retrained PAF still cannot obtain accurate human poses in the challenging CUHK-PEDES dataset due to the occlusion and lighting change. The upper right three images in Fig.~\ref{fig:pose} show the cases of partial and complete missing keypoints. Looking in detail at the procedure of pose estimation, we find that the 14 confidence maps prior to keypoints generation can convey more information about the person in the image when the estimated joint keypoints are incorrect or missing. The lower right images in Fig.~\ref{fig:pose} show the superimposed results of the 14 confidence maps which can still provide cues for human body and its parts.

The pose confidence maps play the two-fold role in our model. On one hand, the 14 confidence maps are concatenated with the 3-channel input image to augment visual representation. We extract visual representation using both VGG-16 \cite{Simonyan2014Very} and ResNet-50 \cite{He2015Deep} on the augmented 17-channel input and compare their performance in the experiments. Taking VGG-16 for example, we first resize the input image to 384$\times$128 and obtain the feature map $\phi^{'}(I)\in\mathbf{R}^{12\times4\times 512}$ before the last pooling layer of VGG-16. Then we partition the $\phi^{'}(I)$ into 6 horizontal stripes inspired by \cite{sun2018beyond} and average each stripe along the first dimension. The $\phi^{'}(I)$ is transformed into $\phi(I)\in\mathbf{R}^{6\times4\times 512}$, where $6\times4\times 512$ means there are 24 regions and each region is represented by a 512-dimensional vector.

On the other hand, the 14 confidence maps are used to learn latent semantic alignment between noun phrase and image regions, which is explained in fine-grained alignment network.

\subsection{Textual Representation Learning}
\textbf{Textual Description.} Given a textual description $T$,
we represent its $j$-th word as an one-hot vector $t_{j} \in \mathbf{R}^{K}$ according to the index of this word in the vocabulary, where $K$ is the vocabulary size. Then we embed the one-hot vector into a 300-dimensional embedding vector:
\begin{equation}
    x_{j}=W_{t} t_{j},
\end{equation}
where $W_{t} \in \mathbf{R}^{300\times K}$ is the embedding matrix.
To model the dependencies between adjacent words, we adopt a bi-directional long short-term memory network (bi-LSTM) \cite{hochreiter1997long} to handle the embedding vectors $X = (x_{1}, x_{2}, ..., x_{r})$ which correspond to the $r$ words in the text description:
\begin{equation}
   \overrightarrow{h_{t}}  = \overrightarrow{LSTM}(x_{t}, \overrightarrow{h_{t-1}}),
\end{equation}
\begin{equation}
   \overleftarrow{h_{t}}  =  \overleftarrow{LSTM}(x_{t}, \overleftarrow{h_{t+1}}),
\end{equation}
where $\overrightarrow{LSTM}$ and $\overleftarrow{LSTM}$ represent the forward and backward LSTMs, respectively.

The global textual representation $e^{t}$ is defined as the concatenation of the last hidden states $\overrightarrow{h_{r}}$ and $\overleftarrow{h_{1}}$:
\begin{equation}
   \label{eq:8}
   e^{t} = concat(\overrightarrow{h_{r}}, \overleftarrow{h_{1}})
\end{equation}

\textbf{Noun Phrase.} For the given textual description, we utilize the NLTK \cite{loper2002nltk} to extract the noun phrase $N$.
Similar to textual description, for the $j$-th noun phrase $n_{j}$ in $N = (n_{1}, n_{2}, ..., n_{m})$, we use the words in $n_{j}$ to represent the noun phrase according to Equations 1-4. Therefore, we can obtain the representations of all noun phrases $e^{n} = (e^{n}_{1}, e^{n}_{2}, ..., e^{n}_{m})$. Note that we adopt the same bi-LSTM when encoding the textual description and noun phrase. Moreover, the number of noun phrase $m$ varies in different textual descriptions.

\subsection{Coarse Alignment Network}
Due to the fact that only partial image regions are related to textual description, we propose a coarse alignment network to select the most related image regions by attention mechanism.

First, we transform local visual representation $\phi(I)$ and textual representation $e^{t}$ to the same feature space.
For different horizontal stripes in $\phi(I)=\{\phi^{ij}(I) \in \mathbf{R}^{512} \mid i=1,2, ...,6, j=1,2,3,4\}$, we employ different transformation matrices as follows:
\begin{equation}
   \widetilde{\phi^{ij}(I)} = W_{\phi^i}\phi^{ij}(I),
\end{equation}
\begin{equation}
   \tilde{e^{t}} = W_{e^{t}}e^{t},
\end{equation}
where $W_{\phi^i}\in\mathbf{R}^{b\times 512}$ and $W_{e^{t}}\in\mathbf{R}^{b\times 2d}$ are two transformation matrices, and $b$ is the dimension of the feature space. Here $d$ is the hidden dimension of the bi-LSTM in textual representation learning.

Then, we define the local cosine similarity between each image region representation $\widetilde{\phi^{ij}(I)}$ and textual representation $\tilde{e^{t}}$:
\begin{equation}
    s_{ij} = \cos(\widetilde{\phi^{ij}(I)}, \tilde{e^{t}}).
\end{equation}
Accordingly, a set of local similarities can be obtained.

Rather than summing all the local similarities as the final score, we propose a hard attention to select the description-related image regions and ignore the irrelevant ones.
Specifically, we set a threshold $\tau$ and select the local similarities if their weights are higher than $\tau$.
The final similarity score $S^{ca}$ is defined as follows:
\begin{equation}
   \label{eq:final_sim}
   q_{ij} = \frac{\exp(s_{ij})}{\sum_{i=1}^{6} \sum_{j=1}^{4} \exp(s_{ij})},
\end{equation}
\begin{equation}
   S^{ca} = \sum_{q_{ij} \ge \tau}{s_{ij}},
\end{equation}
where $q_{ij}$ is the weight of local similarity and $\tau$ is set to $\frac{1}{6\times 4}$. Practically, we can select a fixed number of similarity scores according to their values, but the complex person images can easily fail this idea. In the ablation experiments, we compare the proposed hard attention with several other selection strategies. The experimental results demonstrate the effectiveness of our model.

\subsection{Fine-Grained Alignment Network}
In fact, a noun phrase in the textual description usually describes specific part of an image. To extract the phrase-level related visual contents, we propose a fine-grained alignment network to learn the latent semantic alignment between noun phrase and image regions.
Considering that human pose is closely related to human body parts, we propose to use pose information to guide the matching. As described above, for an image, we extract 14 confidence maps which are related to 14 keypoints. As shown in Fig.~\ref{fig:model1}, 14 keypoints are classified into 6 body parts: head, upper torso, arm, hand, leg, foot. It should be noted that keypoints in different parts have overlap.

For each body part, we first add the corresponding confidence maps of keypoints. Next the confidence maps are embedded into a $b$-dimensional vector $p\in\mathbf{R}^{b}$ by a pose CNN, which has 4 convolutional layers and a fully connected layer. Then two pose-guided attention models are employed to concentrate on the noun phrases and image regions, respectively.
The formulation is illustrated in Fig.~\ref{fig:model1}. Taking textual attention for example, each body part $(p_{1}, p_{2}, ..., p_{6})\in P$ attends to noun phrases with respect to the cosine similarities. Specifically, for the head part $p_{1}$, we first transform all noun phrases $(e^{n}_{1}, e^{n}_{2}, ..., e^{n}_{m})$ to the same feature space as $p_{1}$ and then compute the cosine similarity between $p_{1}$ and all transformed noun phrases $(\tilde{e^{n}_{1}}, \tilde{e^{n}_{2}}, ..., \tilde{e^{n}_{m}})$.
The attended head-related textual representation is defined as follows:
\begin{equation}
    e^{pn}_{1} = \sum_{i=1}^{m}\alpha_{1,i}\tilde{e^{n}_{i}},
\end{equation}
\begin{equation}
    \alpha_{1,i} = \frac{\exp(\cos(p_{1}, \tilde{e^{n}_{i}}))}{\sum_{i=1}^{m}\exp(\cos(p_{1}, \tilde{e^{n}_{i}}))},
\end{equation}
where $\alpha_{1,i}$ indicates the weight value of attention.
Similarly, we can obtain the attended part-related visual representations $(\phi^{p}_{1}, \phi^{p}_{2}, ..., \phi^{p}_{6})$. Then the similarity between attended image-text representations is defined as follows:
\begin{equation}
    s_{i} = \cos(e^{pn}_{i}, \phi^{p}_{i}), i=1,2,..,6
\end{equation}
\begin{equation}
    S^{fa} = \sum_{i=1}^{6}s_{i},
\end{equation}
where $s_{i}$ measures the local similarity between corresponding cross-modal parts.

\subsection{Learning PMA}
The ranking loss $L_{r}$ is a common objective function for the retrieval task. In this paper, we employ the triplet ranking loss as proposed in \cite{Faghri2017VSE} to train our PMA. This ranking loss function ensures that the positive pair is closer than the hardest negative pair in a mini-batch with a margin $\alpha$.
For our coarse alignment network, we obtain a ranking loss $L_{r}^{ca}$.

In addition, we adopt the identification loss and classify persons into different groups by their identities, which ensures the identity-level matching. The image and text identification losses $L_{id_{i}}^{ca}$ and $L_{id_t}^{ca}$ in coarse alignment network are defined as follows:
\begin{equation}
    \widetilde{\phi(I)_{avg}} = avgpool(\widetilde{\phi(I)}),
\end{equation}
\begin{equation}
    L_{id_{i}}^{ca} = -y_{id}\log(softmax(W_{id}^{ca}\widetilde{\phi(I)_{avg}})),
\end{equation}
\begin{equation}
    L_{id_t}^{ca} = -y_{id}\log(softmax(W_{id}^{ca}\tilde{e^{t}})),
\end{equation}
where $W_{id}^{ca}\in\mathbf{R}^{11003\times b}$ is a shared transformation matrix to classify the 11003 different persons in the training set, $y_{id}$ is the ground truth identity. We share $W_{id}^{ca}$ between image and text to constrain their representations in the same feature space.

Then the total loss in CA is defined as:
\begin{equation}
    L^{ca} = L_{r}^{ca}+\lambda_1 L_{id_{i}}^{ca} + \lambda_2 L_{id_t}^{ca},
\end{equation}
where $\lambda_{1}$ and $ \lambda_{2}$ are empirically set to 1 in our experiments. Similarly, we can obtain the total loss $L^{fa}$ in FA .

To make sure that different pose parts can attend to different parts of image and text in fine-grained alignment network, we add an additional part loss $L_{p}$ to classify 6 pose parts into 6 kinds. The part loss $L_{p}$ is defined as follows:
\begin{equation}
    L_{p} = \frac{1}{6}\sum_{i=1}^{6}-y_{p_{i}}\log(softmax(W_{p}p_{i})),
\end{equation}
where $W_{p} \in \mathbf{R}^{6\times b}$ is a transformation matrix and $y_{p_{i}}$ is the ground truth label of $p_{i}$.

Finally, the total loss is defined as:
\begin{equation}
    L = L^{ca}+\lambda_3 L^{fa} + \lambda_{4} L_{p},
\end{equation}
where $\lambda_{3}$ and $\lambda_{4}$ are also empirically set to 1 in our experiments.

During testing, we rank the similarity score $S=S^{ca}+ \lambda_3S^{fa}$ to retrieve the person images based on the text query.

\section{Experiments}
In this section, we first introduce the experimental setup including dataset, evaluation metrics, and implementation details. Then, we analyze the quantitative results of our method and a set of baseline variants. Finally, we visualize several attention maps.

\subsection{Experimental Setup}

\textbf{Dataset and Metrics.} The CUHK-PEDES is currently the only dataset for text-based person search.
We follow the same data split as \cite{Li2017Person}. The training set has 34054 images, 11003 persons and 68126 textual descriptions. The validation set has 3078 images, 1000 persons and 6158 textual descriptions. The test set has 3074 images, 1000 persons and 6156 textual descriptions. On average, each image contains 2 different textual descriptions and each textual description contains more than 23 words. The dataset contains 9408 different words. In addition, the method we proposed can be utilized in other human related cross-modal tasks.

We adopt top-1, top-5 and top-10 accuracies to evaluate the performance.
Given a textual description, we rank all test images by their similarities with the query text. If top-k images contain
any corresponding person, the search is successful.

\textbf{Implementation Details.} In our experiments, we set both the hidden dimension of bi-LSTM and dimension $b$ of the feature space as 1024. For pose CNN, the kernel size of each convolutional layer is $3\times3$ and the numbers of the convolutional channels are 64, 128, 256 and 256, respectively. The fully connected layer has 1024 nodes. In addition, the pose CNN is randomly initialized.
After dropping the words that occur less than twice, the obtained vocabulary has 4984 words.

We initialize the weights of visual CNN with VGG-16 or ResNet-50 pre-trained on the ImageNet classification task. In order to match the dimension of the augmented first layer, we directly copy the averaged weight along the channel dimension to initialize the first layer.
To better train our model, we divide the model training into two steps. First, we fix the parameters of pre-trained visual CNN and only train the other model parameters with a learning rate of $1e^{-3}$. Second, we release the parts of the visual CNN and train the entire model with a learning rate of $2e^{-4}$.
We stop training when the loss converges.
The model is optimized with the Adam \cite{Kingma2014Adam} optimizer. The batch size and margin are 128 and 0.2, respectively.

\begin{table}[t]
\caption{Comparison with the state-of-the-art methods using the same visual CNN as us on CUHK-PEDES. Top-1, top-5 and top-10 accuracies (\%) are reported. The best performance is bold.}
\centering
\label{table:results1}
\resizebox{0.95\columnwidth}!{
\begin{tabular}{l|l|lcl}
\hline
Method&Visual&Top-1&Top-5&Top-10 \\
\hline
{CNN-RNN\shortcite{Reed2016Learning}}&VGG-16&8.07&{-}&32.47\\
{Neural Talk\shortcite{Vinyals2015Show}}&VGG-16&13.66&{-}&41.72\\
{GNA-RNN\shortcite{Li2017Person}}&VGG-16&19.05&{-}&53.64\\
{IATV\shortcite{Li2017Identity}}&VGG-16&25.94&{-}&60.48\\
{PWM-ATH\shortcite{Chen2018Improving}}&VGG-16&27.14&{49.45}&61.02\\
{Dual Path\shortcite{zheng2017dual}}&VGG-16&32.15&{54.42}&64.30\\
\textbf{PMA(ours)}&\textbf{VGG-16}&\textbf{47.02}&\textbf{68.54}&\textbf{78.06}\\
\hline
\hline
{Dual Path\shortcite{zheng2017dual}}&Res-50&44.40&66.26&75.07\\
{GLA\shortcite{chen2018improve}}&Res-50&43.58&66.93&76.26\\
\textbf{PMA(ours)}&\textbf{Res-50}&\textbf{53.81}&\textbf{73.54}&\textbf{81.23}\\

\hline
\end{tabular}
}
\end{table}

\subsection{Quantitative Results}

\textbf{Comparison with the State-of-the-art Methods.} Table~\ref{table:results1} shows the comparison results with the state-of-the-art methods which use the same visual CNN (VGG-16 or ResNet-50) as we do.
Overall, it can be seen that our PMA achieves the best performances under top-1, top-5 and top-10 metrics. Specifically, compared with the best competitor Dual Path \cite{zheng2017dual}, our PMA significantly outperforms it under top-1 metric by about 15\% with the VGG-16 feature and 9\% with the ResNet-50 feature, respectively. The improved performances over the best competitor indicate that our PMA is very effective for this task. Compared with the methods (GNA-RNN \cite{Li2017Person}, IATV \cite{Li2017Identity}, PWM-ATH \cite{Chen2018Improving} and GLA \cite{chen2018improve}) which employ the attention mechanism to extract visual representations or textual representations, our PMA also achieves better performances under three evaluation metrics, which proves the superiorities of our similarity-based attention and pose-guided attention in selecting multi-granularity image-text relations so as to learn diverse and discriminative representations.
Although GLA \cite{chen2018improve} also learns the cross-modal representations by global and local
associations, the improved performance (10\%) over it suggests that our pose-guided multi-granularity attention network can
learn more discriminative and robust multi-granularity representations.

\begin{table}[t]
\caption{Effects of different vocabulary sizes and LSTMs on CUHK-PEDES. ResNet-50 is utilized as the visual CNN.}
\centering
\label{table:results4}
\begin{tabular}{l|lll}
\hline
Method&Top-1&Top-5&Top-10 \\
\hline
{Base-9408}&46.51&68.45&77.32\\
{Base-LSTM}&47.02&69.70&78.33\\
{Base(bi-LSTM)}&\textbf{48.67}&\textbf{70.53}&\textbf{79.22}\\
\hline
\end{tabular}
\end{table}

\begin{table}[t]
\caption{Ablation analysis. We investigate the effectiveness of the concatenated pose before visual CNN (Con-pose), coarse alignment network (CA), and fine-grained alignment network (FA).}
\centering
\label{table:results-abla}
\begin{tabular}{cll|lll}
\hline
Con-pose&CA&FA&Top-1&Top-5&Top-10 \\
\hline
{$\times$}&{$\times$}&{$\times$}&48.67&70.53&79.22\\
{$\surd$}&{$\times$}&{$\times$}&50.06 &71.27 &79.84\\
{$\surd$ }&{$\surd$ }&{$\times$}&52.20& 72.41 &80.48\\
{ $\surd$}&{$\surd$ }&{$\surd$ }&\textbf{53.81}&\textbf{73.54}&\textbf{81.23}\\
\hline
\end{tabular}
\end{table}

\textbf{Ablation Experiments.} To investigate the several components in PMA, we perform a set of ablation studies. The ResNet-50 is employed as the visual CNN in experiments.

We first investigate the importance of vocabulary size by utilizing all the words in the dataset (9408). The baseline model indicates utilizing the global visual and textual features to compute the similarity in experiments. In addition, there is no pose information.
As shown in Table~\ref{table:results4}, utilizing all the words in the dataset has a negative effect on the accuracy, which demonstrates that low-frequency words could make noise to the model. Then we investigate the effectiveness of bi-LSTM. Compared with unidirectional LSTM, the increased performances illustrate that bi-LSTM is more effective to encode textual description.

Table~\ref{table:results-abla} illustrates the effectiveness of concatenated pose confidence maps before visual CNN (Con-pose), coarse alignment network (CA), and fine-grained alignment network (FA).
The improved performances compared with the strong baseline demonstrate that Con-pose, CA and FA are all effective for text-based person search.
Specifically, concatenating the pose confidence maps with original input image indeed improves the matching performance, which demonstrates that pose information is effective in learning discriminative human-related representation. The performance improvement of Con-pose+CA over Con-pose
by 2.2\% in terms of top-1 metric indicates that CA can help our model select coarse description-related image regions and thus benefit the performances. In addition, the Con-pose+CA+FA outperforms the Con-pose+CA by 1.6\% in terms of top-1 metric, which proves the effectiveness of pose-guided fine-grained aligned matching. The overall performances further indicate that exploiting multi-granularity cross-modal relations are effective in image-text matching by learning sufficient and diverse discriminative representations.

\begin{table}[t]
\caption{Effects of selecting different numbers of regions and utilizing different attention mechanisms in CA. Adaptive hard attention is our proposed similarity-based hard attention model.}
\centering
\label{table:results3}
\resizebox{0.9\columnwidth}!{
\begin{tabular}{l|c|lll}
\hline
Method&Number of&Top-1&Top-5&Top-10 \\
{}&Regions&{}&{}&{} \\
\hline
{PMA(CA-Hard)}&5&51.24&72.05&79.87\\
{PMA(CA-Hard)}&10&53.56&73.16&80.97\\
{PMA(CA-Hard)}&20&53.21&72.98&80.78\\
{PMA(CA-Soft)}&24&52.35&72.66&80.24\\
{PMA(CA-Hard)}&Adaptive&\textbf{53.81}&\textbf{73.54}&\textbf{81.23}\\
\hline
\end{tabular}
}
\end{table}

\begin{figure}[t]
\centering
\includegraphics[width=0.9\columnwidth]{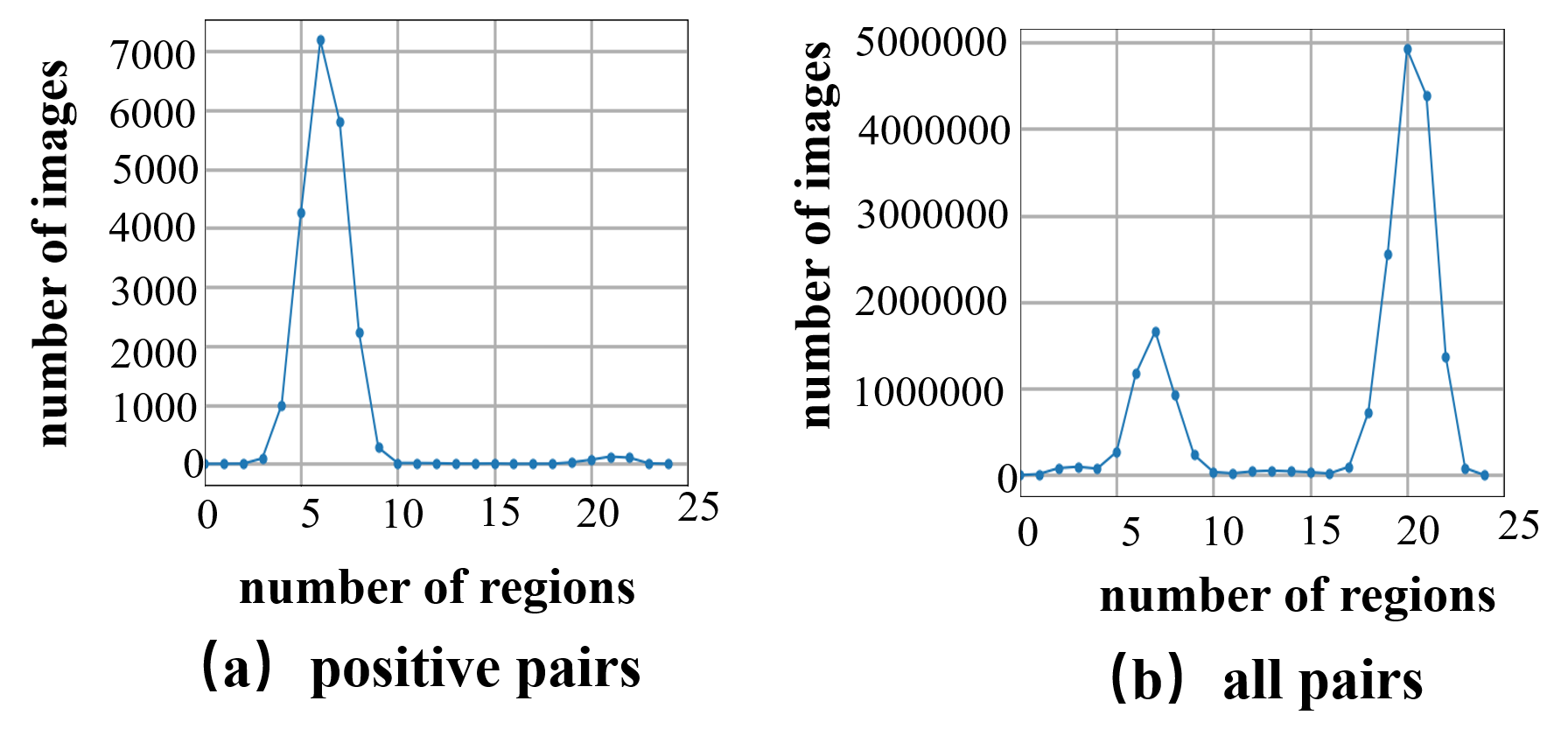}
\caption{The distribution diagram of the selected number of regions by our similarity-based hard attention network. (a) is the distribution diagram of positive pairs. (b) is the distribution diagram of positive and negative pairs.}
\label{fig:plot}
\end{figure}

\begin{figure*}[t]
\centering
\includegraphics[width=0.95\textwidth]{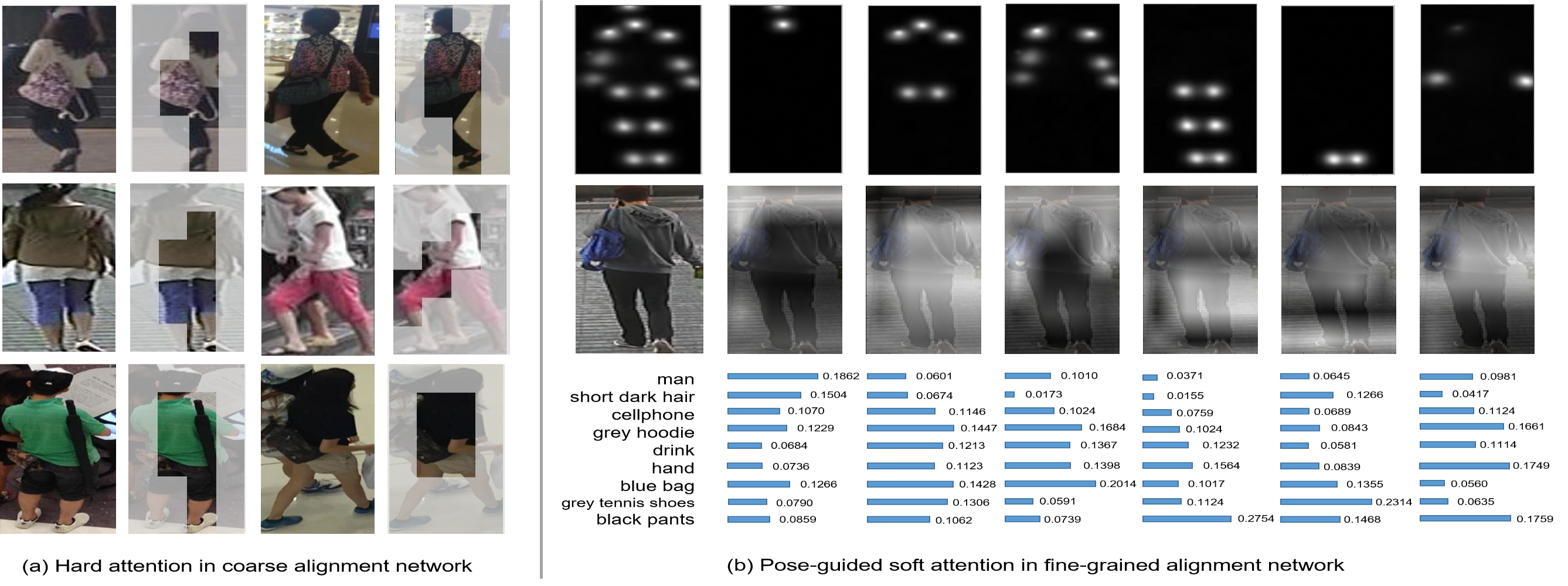}
\caption{Visualization of the similarity-based hard attention regions and pose-guided soft attention map on serval examples. (a) shows the selected regions in coarse alignment network. (b) shows the attention map in fine-grained alignment network.
The lighter regions and higher values indicate the attended areas.}
\label{fig:pose-atten}
\end{figure*}

\textbf{Analysis of Coarse Alignment Network.} In our experiments, we select a variational number of regions by our hard attention. As illustrated in Fig.~\ref{fig:plot}, our model mainly selects 6 regions with positive image-text pairs and 20 regions with negative image-text pairs, which demonstrates that our model can match the positive pairs better. For positive pairs, our PMA can select significant description-related regions. But for negative pairs, due to the lack of regions corresponding to the textual description, the similarity-based hard attention network is unable to select the regions with high similarity score and thus tends to select all regions in image.

We also exploit how the number of selected regions affects the performance of our PMA.
In experiments, we select the top $m$ regions according to their local similarities. Table~\ref{table:results3} shows the experimental results. We can see that the performance increases when increasing the number of selected regions, and saturates soon. It denotes that only some description-related regions are useful for matching. In addition, we compare our hard attention with the soft attention model which re-weights the local similarities according to their values. From the results, we can see that the soft attention performs worse than most of the hard attention, which demonstrates that our hard attention in CA is more effective due to filtering out unrelated visual features.

\begin{table}[t]
\caption{Effects of pose information and noun phrases used in fine-grained alignment network.}
\centering
\label{table:resultsFA}
\begin{tabular}{c|lll}
\hline
Method&Top-1&Top-5&Top-10 \\
\hline
{PMA (FA w/o pose)}&52.54&72.69&80.73\\
{PMA (FA w/o noun phrase)}&52.71&72.87&80.69\\
{PMA (FA)}&\textbf{53.81}&\textbf{73.54}&\textbf{81.23}\\
\hline
\end{tabular}
\end{table}

\textbf{Analysis of Fine-Grained Alignment Network.} In experiments, we utilize 6 body parts rather than 14 score maps of body joints to guide the attention due to the fact that 14 score maps of body joints are just the coordinates of the joints. Therefore, it is difficult to model the relationships with image regions, as well as with the text. To investigate the effectiveness of pose information and noun phrases used in fine-grained alignment network, we perform ablation analysis by dropping one of them. When dropping the pose information, we utilize noun phrases to guide the attention of image regions by their similarities. In addition, dropping the noun phrases means we utilize every word in sentence. As shown in Table~\ref{table:resultsFA}, the performance decreases when dropping one of the key components in FA, which indicates that pose information is effective in selecting aligned image-text contents and noun phrase has more complete semantic information than word. Moreover, we perform the FA with hard attention just like CA but do not obtain meaningful results.

\subsection{Qualitative Results}
To verify whether the proposed model can selectively attend to the corresponding regions and make our matching procedure more interpretable, we visualize the attention weights of the similarity-based hard attention model and pose-guided soft attention model, respectively. Fig.~\ref{fig:pose-atten} shows the results, where lighter regions and higher values indicate the attended areas. We can see that the similarity-based hard attention indeed selects the image regions about the described human and filters out the unrelated regions. The pose-guided soft attention model can attend to the corresponding image regions and noun phrases to pose parts, which illustrates our model can learn accurate aligned fine-grained matching
by the guidance of pose information. Specifically, for the foot part, the model mainly attends to the foot region in image and grey tennis shoes in noun phrases. This fine-grained matching benefits the inference of image-text similarity.

\section{Conclusion}
We propose a novel pose-guided multi-granularity attention network for text-based person search in this paper.
Both coarse alignment network and fine-grained alignment network are proposed to learn multi-granularity cross-modal relevances. The coarse alignment network selects the description-related image regions by a similarity-based attention, while the fine-grained alignment network employs pose information to guide the attention of phrase-related visual contents. Extensive experiments with ablation analysis on a challenging dataset show that our approach outperforms the state-of-the-art methods by a large margin.

\section{Acknowledgments}
This work is jointly supported by National Key Research and Development Program of China (2016YFB1001000), National Natural Science Foundation of China (61525306, 61633021, 61721004, 61420106015, 61572504), Capital Science and Technology Leading Talent Training Project (Z181100006318030), and Beijing Science and Technology Project (Z181100008918010). This work is also supported by grants from NVIDIA and the NVIDIA DGX-1 AI Supercomputer.

\bibliographystyle{aaai}
\bibliography{138.egbib}

\end{document}